\documentclass[nohyperref]{article}
\usepackage{microtype}
\usepackage{graphicx}
\usepackage{subfigure}
\usepackage{booktabs} 
\usepackage{hyperref}

\usepackage[accepted]{icml2023}

\usepackage{times}
\usepackage{latexsym}
\usepackage{hyperref}
\usepackage{url}
\usepackage{bbm}
\usepackage{amsmath,amsfonts,bm}
\usepackage{graphicx}
\usepackage{multirow}
\usepackage{makecell}
\usepackage{array}
\usepackage[para]{threeparttable}
\usepackage{booktabs}
\usepackage{tabularx}
\usepackage{xspace}
\usepackage{multirow}
\usepackage[normalem]{ulem}
\usepackage{longtable}
\usepackage{kotex}

\usepackage[capitalize,noabbrev]{cleveref}

\usepackage[textsize=tiny]{todonotes}

\icmltitlerunning{Exploring the Benefits of Training Expert Language Models over Instruction Tuning}

\begin{document}

\twocolumn[
\icmltitle{Exploring the Benefits of Training Expert Language Models\\ over Instruction Tuning}



\icmlsetsymbol{equal}{*}

\begin{icmlauthorlist}
\icmlauthor{Joel Jang}{school,comp}
\icmlauthor{Seungone Kim}{school}
\icmlauthor{Seonghyeon Ye}{school,comp}
\icmlauthor{Doyoung Kim}{school}
\icmlauthor{Lajanugen Logeswaran}{comp}
\icmlauthor{Moontae Lee}{comp,school2}
\icmlauthor{Kyungjae Lee}{comp}
\icmlauthor{Minjoon Seo}{school}
\end{icmlauthorlist}

\icmlaffiliation{school}{KAIST}
\icmlaffiliation{comp}{LG AI Research}
\icmlaffiliation{school2}{University of Illinois Chicago}

\icmlcorrespondingauthor{Joel Jang}{joeljang@kaist.ac.kr}

\icmlkeywords{Machine Learning, ICML}

\vskip 0.3in
]



\printAffiliationsAndNotice{Work done while JJ and SY were interning at LG AI Research.}

\begin{abstract}
Recently, Language Models (LMs) instruction-tuned on multiple tasks, also known as multitask-prompted fine-tuning (MT), have shown the capability to generalize to unseen tasks. Previous work has shown that scaling the number of training tasks is the key component in making stronger MT LMs. In this work, we report an unexpected finding that an \textit{expert} LM fine-tuned on just a single task can outperform an MT LM trained with 300+ different tasks on 11 different unseen datasets and on 13 datasets of the BIG-bench benchmark by a mean accuracy of 3.20\% and 1.29\%, respectively. This finding casts doubt on the previously held belief that simply scaling the number of tasks makes stronger MT LMs. Leveraging this finding, we further show that this distributed approach of training a separate expert LM per training task instead of a single MT LM for zero-shot inference possesses many benefits including (1) avoiding negative task transfer that often occurs during instruction tuning, (2) being able to continually learn new tasks without having to re-train on previous tasks to avoid catastrophic forgetting, and (3) showing \textit{compositional} capabilities when merging individual experts together. The code is available at \href{https://github.com/joeljang/ELM}{https://github.com/joeljang/ELM}.
\end{abstract}

\vspace{-.5em}

\section{Introduction}
\label{introduction}
\begin{figure}[t!]
    \centering
    \includegraphics[width=1\linewidth]{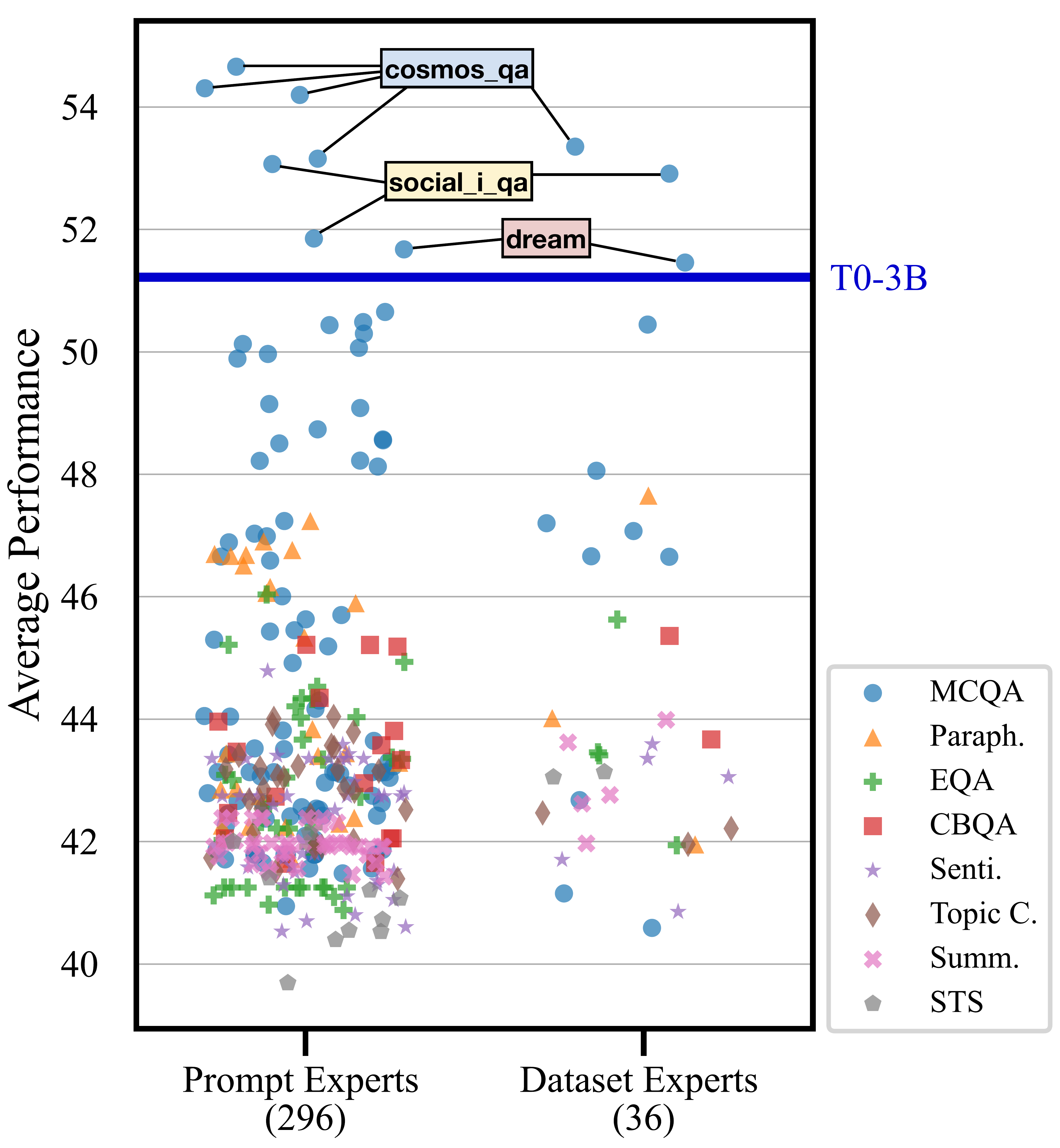}
    \caption{\footnotesize Mean accuracy performance of Expert LMs (each trained on a single task) on 11 unseen datasets compared to an instruction-tuned LM, T0-3B. Results show some Expert LMs surpassing T0-3B, challenging the commonly held belief that simply scaling the total number of training tasks is the key component to enhancing the capability of MT LMs.}
    \label{fig:analysis}
\end{figure} 
Recent works show pretrained Language Models (LMs) that have been fine-tuned on multiple tasks with instructions (prompted instances), also known as multitask-prompted fine-tuned LMs and referred to as MT LMs in this work, can generalize to unseen tasks without task-specific fine-tuning~\citep{wei2021finetuned, sanh2021multitask, chung2022scaling, ye2022guess, ouyang2022training, wang2022super, muennighoff2022crosslingual}. 
This paper raises some questions regarding the current paradigm of training MT LMs and is mainly divided into two parts. In Part 1, we report an unexpected finding regarding \textit{expert} LMs (trained only on a single task) compared to MT LMs. In Part 2, we leverage the finding to highlight some of the benefits of \textit{expert} LMs over MT LMs.

\begin{figure*}[t!]
    \centering
    \includegraphics[width=0.8\linewidth]{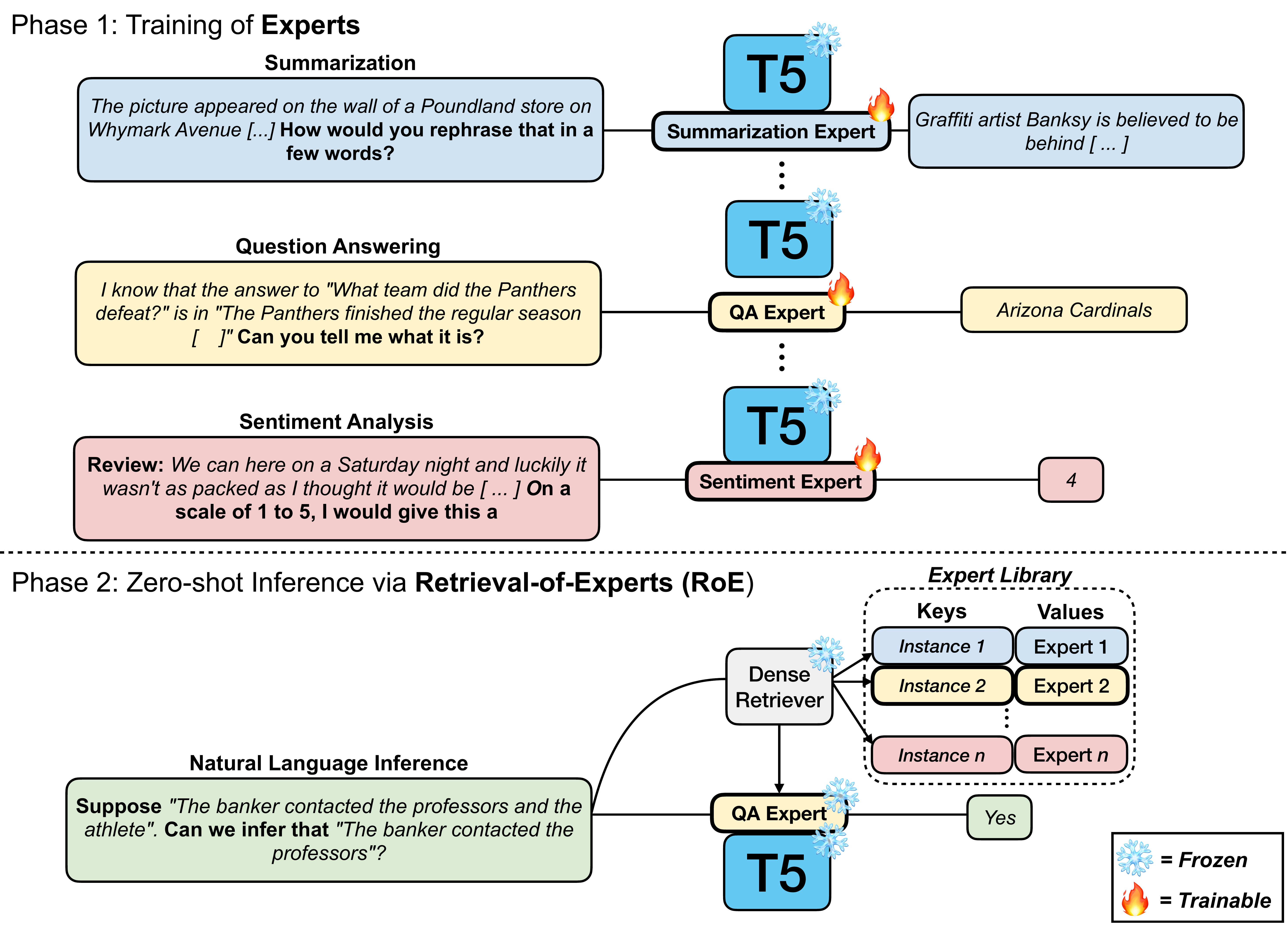}
    \caption{\footnotesize Independent training and Retrieval-of-Experts (RoE) for zero-shot task generalization. During training, only the additional adapters (experts) are trained while the backbone LM is frozen. After training separate experts per training task, we construct an Expert Library that stores samples of the training task as \textit{keys}, and the specific expert id as \textit{values}. During zero-shot inference, the most relevant expert is retrieved for an unseen task.}
    \label{fig:fig1}
\end{figure*} 

\paragraph{Part 1 (Section \ref{sec:experimental_results})} Previously, the key component to enhancing the unseen task generalization performance of MT LMs was thought to be scaling the total number of tasks used in training~\citep{wei2021finetuned, chung2022scaling, wang2022super}. However, in this work, we show that training a single expert LM on \emph{one}\footnote{Training task: cosmos\_qa, Prompt Name: no\_prompt\_text from \citet{bach2022promptsource}.} out of the 300+ tasks used to train an MT LM (T0-3B~\citep{sanh2021multitask}) can outperform the MT LM by a non-trivial margin on 24 unseen tasks on mean accuracy.

Specifically, following the same experimental setup (training and evaluation) as T0-3B~\citep{sanh2021multitask}, one of the most widely used MT LM, we first train \textit{expert} LMs for each given training task (296) by freezing the underlying LM and updating adapters~\citep{pmlr-v97-houlsby19a}. We report a finding that shows 7 out of the 296 experts surpass T0-3B on the capability to generalize to unseen tasks on mean accuracy (shown in Figure \ref{fig:analysis}). Using the top performing expert for all of the unseen task evaluation tasks surpasses T0-3B by a mean accuracy of 3.20\% and 1.29\% on 11 unseen datasets and 13 datasets of the BIG-Bench benchmark, respectively. We also show that applying a simple mechanism to retrieve relevant experts for each individual unseen task results in comparable performance to T0-3B. Considering the significant room for improvement when retrieving the best-performing expert for each unseen task (+11.94\% compared to T0-3B), these results imply that choosing the right expert rather than naïvely utilizing a single MT LM for all of the unseen tasks can be a more efficient and effective approach. 

\paragraph{Part 2 (Section \ref{sec:experiment_part2})} Leveraging the finding of expert LMs showing improved unseen task generalization capability, we highlight three other advantages of training multiple expert LMs for each task and retrieving the relevant expert during inference (shown in Figure \ref{fig:fig1}) compared to training MT LMs.

\textbf{\#1.} MT LMs do not show the optimal performance for \textit{seen} tasks because of negative task transfer, where learning multiple tasks at once hinders the learning of some specific tasks~\citep{aghajanyan-etal-2021-muppet, asai2022attempt, zhang2022survey}. Expert LMs, on the other hand, are not subject to negative task transfer~\citep{levine2022standing} since each task is learned independently; We show our approach of selecting relevant experts during inference results in a +10.4\% mean accuracy improvement on validation datasets of the 36 training tasks compared to T0-3B.

\textbf{\#2.} MT LMs are susceptible to catastrophic forgetting~\citep{mccloskey1989catastrophic} of previous tasks when learning new tasks and require re-training on previous tasks to mitigate forgetting~\citep{chakrabarty2022finetuned}. Results show our \textit{distributed} (training individual tasks in an independent manner) approach results in absolutely no degradation of seen tasks, even when adding the 8 new experts to the Expert Library, without re-training on previous tasks when learning 8 new generative tasks. 

\textbf{\#3.} We show that MT LMs show poor ability in performing \textit{composition} of previously learned tasks given via concatenation of the corresponding instructions as a single \textit{compositional} instruction. On the other hand, we show that \textit{merging} the two experts trained on the individual tasks with mT5-3B~\citep{xue-etal-2021-mt5} as the underlying pre-trained LM results in an expert that can outperform its MT LM counterpart, mT0-3B~\citep{muennighoff2022crosslingual}, by a mean ROUGE-L score of +2.71 on 5 novel compositional tasks (summarization \& translation). Details of the merging mechanism are provided in Section \ref{method:merge}.

\section{Related Work}
\label{related_work}
\subsection{Multitask Prompted Fine-tuning of Language Models}
Several studies have demonstrated that multitask fine-tuning moderately sized LMs with instructions, also referred to as \textit{instruction tuning}, enables zero-shot task generalization. Specifically, \citet{sanh2021multitask, wang2022super} have shown that scaling the number of training tasks, the number of prompts per task, and the size of the LM helps boost zero-shot task generalization performance. In addition to scaling these aspects, \citet{chung2022scaling} include Chain-of-Thought~\citep{wei2022chain} tasks during instruction tuning, reaching state-of-the-art performance on zero-shot and few-shot settings with PaLM 540B~\citep{chowdhery2022palm} as the underlying LM. \citet{lin2022unsupervised} improve MT LMs by adapting MT LMs on subsets of the training data retrieved given a few unlabeled examples of the unseen task. \citet{ouyang2022training} adapt MT LMs to align with human preferences through reinforcement learning. \citet{muennighoff2022crosslingual} include multilingual tasks to show cross-lingual generalization capability. \citet{ye2022guess} flip the instruction and label space to enhance generalization capability to novel unseen labels. \citet{asai2022task} utilize instruction tuning to construct a general-purpose retrieval system. Similarly, \citet{su2022one} utilize instruction tuning to construct a general-purpose embedding model that can be used to perform different unseen tasks requiring text embeddings. 

While previous literature has mostly asserted that the primary key component of MT LMs is scaling the total number of training tasks, in this paper, we propose an alternative perspective and instead show experimental results and findings that the \textit{feature} of the tasks may be a more critical factor (analysis provided in Section \ref{sec:experimental_results}); Similar findings are shown in the setting of few-shot adaptation~\citep{chan2022few} as well. 

\subsection{Retrieving task-specific embeddings}
Retrieving task-specific parameters has the advantage of rapid target task adaptation, especially for low-resource scenarios \citep{vu-etal-2022-spot, asai2022attempt, ye2022retrieval, qin-Eisner-2021-learning, wang2022learning, bari2022spt}. \citet{vu-etal-2022-spot} show that retrieving an optimal source soft prompt leads to better initialization for adapting to the target task. \citet{asai2022attempt} also focus on retrieval of soft prompts for initialization for the target task but utilize the idea of attention weights to effectively interpolate between multiple training soft prompts. Similarly, \citet{ye2022retrieval} extend this idea of retrieving soft prompts, but utilize an MT LM as the underlying LM and do not fine-tune the LM to the target task, performing the target task in a zero-shot manner. Our work is motivated by \citet{ye2022retrieval}, but proposes to replace the instruction tuning stage altogether, using vanilla pretrained LMs as the underlying LM instead of MT LMs. We accomplish this by training experts whereas previous work trained soft prompts on top of MT LMs.

\subsection{Distributed Training of Language Models}
Recent work has shown the possibilities and benefits of distributed training of LMs. \citet{li2022branch} have shown that it is possible to merge individual LMs pretrained on different subsets (domains) of the training corpora to construct a single LM that shows lower overall perplexity compared to an LM trained on all of the corpora at once. Another line of work that explores merging individually fine-tuned LMs is \citet{wortsman2022model}, where they merge LMs fine-tuned on the same task with different configurations to boost performance. Similarly, \citet{wortsman2022fi} merge LMs fine-tuned on the same task, but with subsets of the training data for efficiency. \citet{don2022cold} explore merging LMs fine-tuned on different tasks to make a multitask fine-tuned LM in a distributed manner, which has many benefits including federated learning~\citep{mcmahan2017communication}. 

Other interesting extensions of distributed LM training include performing task arithmetic with task vectors~\citep{ilharco2022editing}, training and performing inference of several billion parameter LMs on distributed compute~\citep{borzunov2022petals}, introducing language-specific modules for growing the total capacity of multilingual LMs~\citep{pfeiffer2022lifting}, finding theoretical guarantees of why merging works~\citep{frankle2020linear, ainsworth2022git} and proposing novel methods of merging model weights~\citep{matena2021merging}. In our work, we also show the benefits of distributed LM training by showing that the capability of expert LMs can be further amplified through \textit{merging} individual experts.

\section{Expert Language Models}
In this section, we describe the framework of our proposed method. We train each expert by training adapters for each training task (Section \ref{method:training}). During inference, we retrieve the relevant experts from the Expert Library (Section \ref{method:roe}). We additionally explore the effect of merging experts to observe the benefits of distributed training (Section \ref{method:merge}).

\subsection{Training Experts}
\label{method:training}
\begin{figure}[t!]
    \centering
	\includegraphics[width=0.47\textwidth]{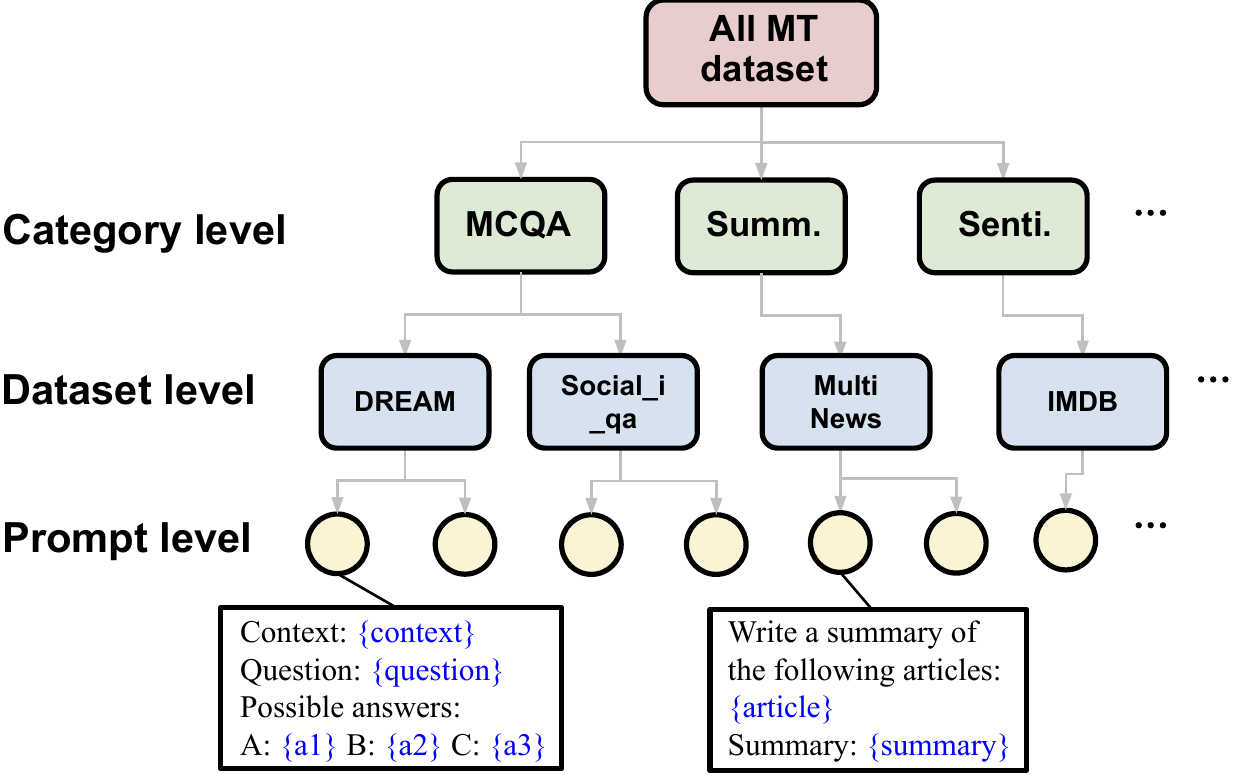}
    \caption{\footnotesize The hierarchy of all the training datasets used to train MT LMs. In this work, we explore training Dataset level Experts (DE) and Prompt level Experts (PE).}
    \label{fig:hierarchy}
\end{figure}
For training the experts, we mainly explore parameter-efficient fine-tuning via adapters while freezing the underlying LM to train individual experts. We train experts for each task with the corresponding \textit{prompts} and denote the resulting experts as Prompt Experts (PE).\footnote{Each prompt (instruction) is referred to as \textit{tasks}, following \citet{chung2022scaling}.} We also explore training experts for each \textit{dataset}, which consists of multiple training prompts, referred to as Dataset Experts (DE). For training DE, instead of utilizing a parameter-efficient fine-tuning approach (adapters), we instead simply train the entire LM to observe the merging capability of expert LMs.\footnote{Experimental results show that merging adapter experts does not lead to improved positive task transfer on mean accuracy (shown in Section \ref{sec:experimental_results}).} 
Figure \ref{fig:hierarchy} shows the hierarchy of the training datasets and the level at which PE and DE are trained on.


\paragraph{Adapters} We apply a parameter-efficient method of representing experts by training additional adapters while freezing the original parameters~\citep{pmlr-v97-houlsby19a}. Specifically, given a standard Transformer LM with $l$ layers, input sequence $X$  containing $T$ tokens, the output for a single layer $\textbf{h}^l_{1:T}$ is calculated by
\begin{equation}\label{eq1}
\textbf{h}^l_{t} = \textsc{FFN}_{d}(\textbf{u}^l_{t}) + \textbf{u}^l_{t},
\end{equation}
\begin{equation}\label{eq2}
\textbf{u}^l_{1:T} = \textsc{Self-Att}(\textbf{h}^{l-1}_{1:T}) + \textbf{h}^{l-1}_{1:T},
\end{equation}
where $\textbf{h}^l_{t}$ is the hidden state of \textit{t}-token after the \textit{l}-th layer, \textsc{Self-Att}(·) is the self-attention module, and $\textsc{FFN}_{d}$(·) is the feed-forward network with hidden dimensions \textit{d}. When fine-tuning the LM with an adapter expert, each layer before the self-attention layer (Equation \ref{eq1}) changes into the following format:
\begin{equation}\label{eq3}
\textbf{h}^l_{t} = \textsc{FFN}_e(\textbf{u}^l_{t}) + \textsc{FFN}_{d}(\textbf{u}^l_{t}) + \textbf{u}^l_{t},
\end{equation}
where \textit{e} represents the hidden dimension of the adapter feed-forward network. When using adapters to represent experts, parameters of $\textsc{FFN}_{e}$ are the only trainable parameters and the rest of the parameters in the LM are frozen.

\subsection{Retrieval-of-Experts (RoE)}
\label{method:roe}
After independent (distributed) training of individual experts, we retrieve one of the experts to use during inference~\citep{ye2022retrieval}. To this end, we construct an \textit{Expert Library} and use dense retrieval  to retrieve a relevant expert from the library to use during inference.

\paragraph{Expert Library}
We first construct the \textit{Expert Library}. This library contains keys that are each embedding representations of a single instance from the training tasks and values that are unique ids of the corresponding trained experts. For each unique expert, \textit{S} training instances are randomly sampled and stored in the library. This results in [$S\times\textit{\# of experts}$] entries in the Expert Library. To get the embedding representation of the training instances, we employ a simple Sentence Transformer~\citep{reimers-2019-sentence-bert} as the dense retriever.\footnote{We explore other text embedding models for the retriever such as Sentence-T5, SimCSE, INSTRUCTOR, etc., in Appendix \ref{appen:text_format}. Sentence Transformer shows the best performance among the embedding models.} For the text format of the training instance that is given to the embedding model as input, we simply concatenate the answer choice (e.g. Yes$|$No, A$|$B$|$C$|$D) to the Prompted Input. The answer choice for generative tasks is given as `None'. We report ablation results of varying the text format given as the input to the embedding model in Appendix \ref{appen:text_format}. 

\paragraph{Retrieval}
Following the approach of \citet{lin2022unsupervised, ye2022retrieval}, given a target task during inference, we first randomly select $Q$ instances from the target task\footnote{We assume a scenario where we can perform batch-inference.}. Next, we use the same text format (concatenation of Prompted Input and Answer Choice) and the same embedding model used to construct the Expert Library to obtain embedding representations of each of the $Q$ target queries. We then use MIPS (maximum inner product search) on our Expert Library to identify the most similar training instance (key) for each query instance, resulting in a total of $Q$ corresponding experts (value). We select the most frequently retrieved expert as the expert for solving the given target task. 

\subsection{Merging of Experts}
\label{method:merge}
Previous work has shown the possibility of distributed multitask fine-tuning by \textit{merging} individually fine-tuned LMs~\citep{don2022cold}. Along with selecting the most retrieved expert, we observe how merging fully fine-tuned LMs (DE) affects the generalization performance on the unseen tasks. 

A fully fine-tuned LM can be represented in the form of a vector $\tau_{d} = \theta_{d} - \theta_{pre}$ where $\theta_{pre}$ represent the full parameters of the vanilla pretrained LM and $\theta_{d}$ represents the full parameters of the LM fine-tuned on the training dataset $d$~\citep{ilharco2022editing}. The formula for merging of \textit{N} experts can be denoted as follows:

\begin{equation}\label{eq4}
\theta_{new} = \theta_{pre} + (\sum_{i}^{N}\lambda_{i}\tau_{i})
\end{equation}
where $\lambda_{i} = \frac{1}{N}$ as default if not stated otherwise. Note that when $\lambda_{i} = \frac{1}{N}$, it results in merging experts uniformly. In some cases, however, performance was optimal when $\sum_{i}\lambda_{i} > 1$ and each $\lambda_{i}$ (representing the importance to place on $\tau_{i}$) and was set to a different value determined using a held-out validation dataset following \citet{ilharco2022editing}. A concrete example is provided in Appendix \ref{appen:composition_details}.

\section{Experimental Setup}
\label{sec:experimental_setup}
\begin{table*}[t!]
\fontsize{12}{14}\selectfont
\centering
\resizebox{\textwidth}{!}{\begin{tabular}{lccccccccccc|c}
    \toprule
    \multicolumn{1}{c}{\multirow{2}{*}{\textbf{Method}}}& \multicolumn{5}{c}{NLI} & \multicolumn{3}{c}{Sentence Completion}          & \multicolumn{2}{c}{Coreference Resolut.} & WSD & \multicolumn{1}{c}{\multirow{2}{*}{\textbf{Total Avg.}}} \\ 
    \cmidrule(lr){2-6} \cmidrule(lr){7-9} \cmidrule(lr){10-11} \cmidrule(lr){12-12} & \textbf{RTE} & \textbf{CB} & \textbf{AN. R1} & \textbf{AN. R2} & \textbf{AN. R3} & \textbf{COPA} & \textbf{Hellasw.} & \textbf{StoryC.} & \textbf{Winogr.} & \textbf{WSC} & \textbf{WiC} \\ 
    \midrule
    \textsc{T0-11B} & 80.83 & 70.12 & 43.56 & 38.68 & 41.26 & 90.02 & 33.58 & 92.40 & 59.94 & 61.45 & 56.58 & 60.76\\
    \textsc{GPT-3(175B)} & 63.50 & 46.40 & 34.60 & 35.40 & 34.50 & 91.00 & 78.90 & 83.20 & 70.20 & 65.40 & 45.92 & 59.00\\
    \midrule
    \textsc{T0-3B} & \underline{60.61} & \underline{48.81} & 35.10 & 33.27 & \underline{33.52} & 75.13 & 27.18 & 84.91 & 50.91 & \textbf{65.00} & \underline{51.27} & 51.43 \\
    \textsc{T5(3B) + Cos PE} & 49.53 & \textbf{49.52} & \textbf{36.21} & \textbf{36.11} & \textbf{36.38} & \textbf{89.63} & \textbf{43.77} & \textbf{97.06} & \underline{56.65} & 57.02 & 49.01 & \textbf{54.63} \\
    \textsc{T5(3B) + PE w/ RoE} & \textbf{64.01} & 43.57 & \underline{35.49} & \underline{34.64} & 31.22 & \underline{79.25} & \underline{34.60} & \underline{86.33} & \textbf{61.60} & \underline{62.21} & \textbf{52.97} & \underline{53.48}\\
    \midrule
    \textsc{T5(3B) + PE w/ RoE (Orc.)} & 70.32 & 70.12 & 40.02 & 40.11 & 42.07 & 92.88 & 55.00 & 97.47 & 64.40 & 65.77 & 58.90 & 63.37\\
    \bottomrule        
\end{tabular}}
\caption{\footnotesize Evaluation performance on 11 different unseen datasets categorized into 4 task categories. \textsc{PE} represents Prompt Experts.  \textsc{PE w/ RoE (Orc.)} represents retrieving the best-performing (oracle) expert for each evaluation task. \textsc{Cos PE} represents the PE trained on \textsc{Cosmos-qa} with the prompt \textsc{no-prompt-text} which showed the highest mean accuracy on the 11 unseen tasks. \textsc{PE w/ RoE} represents Retrieval-of-Expert (RoE) for each individual unseen task. Note that PE adds 100M additional parameters while freezing the 3B paramters of T5 during training. The best comparable performances are \textbf{bolded} and second best \underline{underlined}.} 
\label{table:main_unseen}
\vspace{-3mm}
\end{table*}  

\begin{table}[ht!]
\centering
\fontsize{8}{9}\selectfont
\begin{tabular*}{1\columnwidth}{l|cc|ccc}
\toprule
\multirowcell{2}{Dataset (metric)} & \textsc{T0} & \textsc{Cos PE} & \textsc{T0} & \textsc{GPT-3} & \textsc{PaLM} \\
& 3B & 3B & 11B  & 175B & 540B  \\ 
\midrule
Known Un.      & 47.83 & \textbf{58.70} & 65.22 & 60.87 & 56.52\\
Logic Grid     & \textbf{32.10} & 30.70 & 33.67 & 31.20 & 32.10\\
Strategy.      & \textbf{53.23} & 42.36 & 54.67 & 52.30 & 64.00\\
Hindu Kn.      & 34.86 & \textbf{51.43} & 42.86 & 32.57 & 56.00\\
Movie D.       & \textbf{53.22} & 46.72 & 57.33 & 51.40 & 49.10\\
Code D.        & 53.33 & \textbf{66.67} & 51.67 & 31.67 & 25.00\\
Concept        & 67.25 & \textbf{72.92} & 71.72 & 26.78 & 59.26\\
Language       & 14.94 & \textbf{25.95} & 18.33 & 15.90 & 20.10\\
Vitamin        & \textbf{58.18} & 46.55 & 57.33 & 12.30 & 14.10\\
Syllogism      & \textbf{52.27} & 50.00 & 48.33 & 50.50 & 49.90\\
Misconcept.    & \textbf{52.05} & 47.03 & 52.97 & 47.95 & 47.47\\
Logical        & \textbf{45.33} & 42.40 & 54.67 & 23.42 & 24.22\\
Winowhy        & 44.29 & \textbf{44.33} & 55.00 & 51.50 & 45.30\\ 
\midrule
BIG-bench AVG  & 46.84 & \textbf{48.13} & 51.06 & 37.57 & 41.77\\ 
\bottomrule
\end{tabular*}
\caption{\footnotesize Evaluation performance on 13 BIG-bench tasks. The best comparable performances are \textbf{bolded}.}
\vspace{-5mm}
\label{table:big_bench}
\end{table} 

\begin{table*}[ht!]
\fontsize{8}{8}\selectfont
\centering
\begin{tabular}{lcccccccc|c}
\toprule
\multicolumn{1}{c}{\multirow{2}{*}{\textbf{Method}}} & \textbf{wiki auto} & \textbf{HGen} & \textbf{haiku} & \textbf{covid qa} & \textbf{eli5} & \textbf{emdg} & \textbf{esnli} & \textbf{twitter} & \textbf{Total} \\ 
& (BLEU) & (ROUGE) & (ROUGE) & (BS) & (BS) & (BS) & (BS) & (BS) & \textbf{Avg.}\\
\midrule
 \textsc{T0-3B} & \underline{21.76} & \underline{33.29} & 19.93 & \textbf{50.00} & \textbf{59.86} & 47.76 & \underline{42.80} & 28.40 & \underline{37.98} \\
\textsc{T5(3B) + Sam PE}  & \textbf{30.69} & 25.49 & \underline{25.25} & \underline{49.93} & \underline{47.94} & \textbf{51.36} & \textbf{58.28} & \textbf{69.55} & \textbf{44.81}\\
\textsc{T5(3B) + PE w/ RoE} & 3.88 & \textbf{35.55} & \textbf{26.53} & 33.52 & 33.66 & 49.90 & 28.61 & \underline{49.22} & 32.61 \\
\midrule
 \textsc{T5(3B) + PE w/ RoE (Orc.)} & 31.56 & 35.55 & 30.16 & 52.49 & 63.20 & 58.36 & 60.02 & 82.08 & 51.67\\
\bottomrule        
\end{tabular}
\caption{\footnotesize Evaluation performance on 8 unseen generative tasks. \textsc{Sam PE} represents the PE trained on \textsc{Samsum} with the prompt \textsc{Given the above dialogue write a summary} which showed the highest mean score on the 8 unseen generative tasks. The best comparable performances are \textbf{bolded} and second best \underline{underlined}.} 
\label{table:generative}
\vspace{-3mm}
\end{table*} 

\begin{table*}[t!]
\fontsize{12}{14}\selectfont
\resizebox{\textwidth}{!}{\begin{tabular}{lccccccccccc|c}
    \toprule
    \multicolumn{1}{c}{\multirow{2}{*}{\textbf{Method}}}& \multicolumn{5}{c}{NLI} & \multicolumn{3}{c}{Sentence Completion}          & \multicolumn{2}{c}{Coreference Resolut.} & WSD & \multicolumn{1}{c}{\multirow{2}{*}{\textbf{Total Avg.}}} \\ 
    \cmidrule(lr){2-6} \cmidrule(lr){7-9} \cmidrule(lr){10-11} \cmidrule(lr){12-12} & \textbf{RTE} & \textbf{CB} & \textbf{AN. R1} & \textbf{AN. R2} & \textbf{AN. R3} & \textbf{COPA} & \textbf{Hellasw.} & \textbf{StoryC.} & \textbf{Winogr.} & \textbf{WSC} & \textbf{WiC} \\ 
    \midrule
    \textsc{T5(3B) + Cos PE} & \underline{49.53} & \textbf{49.52} & \textbf{36.21} & \textbf{36.11} & \textbf{36.38} & 89.63 & \textbf{43.77} & 97.06 & \textbf{56.65} & \textbf{57.02} & 49.01 & \textbf{54.63} \\
    \textsc{T5(3B) + Soc PE} & \textbf{61.26} & 38.81 & 33.16 & 33.63 & 33.46 & \underline{90.50} & \underline{37.21} & \underline{97.09} & \underline{55.28} & 50.00 & \textbf{50.11} & \underline{52.77} \\
    \textsc{T5(3B) + Cos\&Soc PE (Mer.)} & 49.10 & \underline{39.40} & \underline{33.80} & \underline{34.28} & \underline{34.18} & \textbf{91.63} & 36.29 & \textbf{97.25} & 55.06 & \underline{51.25} & \underline{49.62} & 51.99 \\
    \midrule
    \textsc{T5(3B) + Cos DE} & 59.71 & \textbf{57.62} & 33.45 & 33.93 & 34.54 & \underline{90.00} & \textbf{36.58} & \underline{96.29} & 53.37 & \underline{42.88} & 49.91 & \underline{53.48} \\
    \textsc{T5(3B) + Soc DE} & \textbf{65.52} & 48.69 & \textbf{35.20} & \textbf{35.39} & \textbf{37.11} & 83.25 & 30.38 & 87.18 & \underline{54.27} & \textbf{54.62} & \textbf{51.39} & 53.00 \\
    \textsc{T5(3B) + Cos\&Soc DE (Mer.)} & \underline{60.43} & \underline{54.17} & \underline{35.01} & \underline{34.53} & \underline{35.52} & \textbf{91.25} & \underline{35.59} & \textbf{96.73} & \textbf{54.33} & \underline{42.88} & \underline{50.05} & \textbf{53.68} \\
    \bottomrule        
\end{tabular}}
\caption{\footnotesize Evaluation performance on 11 different unseen datasets categorized into 4 task categories. 
\textsc{PE} represents Prompt Experts. \textsc{Cos PE} represents the PE trained on \textsc{Cosmos-qa} dataset and \textsc{no prompt text} prompt and \textsc{Soc PE} represents the PE trained on \textsc{Social-i-qa} dataset and \textsc{Show choices and generate answer} prompt. \textsc{Cos\&Soc PE (Mer.)} represents expert constructed by performing \textit{uniform} merging with the \textsc{Cos PE} and \textsc{Soc PE}. \textsc{Cos DE} represents the DE trained on the \textsc{Cosmos-qa} dataset with all of the prompts and \textsc{Soc DE} represents the DE trained on \textsc{Social-i-qa} on all of the propmts. \textsc{Cos\&Soc DE (Mer.)} represents expert constructed by performing merging with the \textsc{Cos DE} and \textsc{Soc DE}. The best comparable performances are \textbf{bolded} and second best \underline{underlined}.} 
\label{table:main_merge}
\vspace{-3mm}
\end{table*}  

\paragraph{Training Setup}
Following the setting of \citet{sanh2021multitask}, we use a total of 36 training datasets of T0 for training our experts.\footnote{The original T0~\citep{sanh2021multitask} paper includes 38 training datasets. However, we could not load 4 datasets from the Huggingface Dataset library: adversarial\_qa/dbidaf, adversarial\_qa/dbert, adversarial\_qa/droberta, and duorc/SelfRC. Instead, we utilize the adversarial\_qa/adversarialQA dataset and also additionally train on commonsense\_qa dataset which is a variant of the cos\_e dataset, resulting in a total of 36 training datasets.} For each dataset, we use all of the prompts used to train T0 from the Promptsource Library~\citep{bach2022promptsource} which results in a total of 296 prompts to train the corresponding experts ($\sim$8 prompts per training dataset). This results in 36 Dataset Experts (DE) represented via fully fine-tuned LMs, and 296 Prompt Experts (PE) via adapter training. For each individual fine-tuning, we randomly sample $K=50,000$ training instances for each classification task and $K=10,000$ for each generative task.\footnote{We train with less number of instances for the generative tasks because the training generative tasks required longer max token length, and thus longer training time.} We use the LM-adapted T5 model \cite{lester2021power} checkpoint as our base model, and train for 5 epochs with a constant learning rate of 1e-4 for both adapter fine-tuning and full LM fine-tuning. For the construction of the Expert Library, much smaller $S=100$ training instances are randomly sampled for each expert following \citet{ye2022retrieval}. 

\paragraph{Evaluation Setup} We evaluate the baseline MT LMs (T0-3B, T0-11B) and our proposed method (T5-3B + DE/PE) on the same evaluation setting as the original T0 paper~\citep{sanh2021multitask}: 11 unseen datasets that can be categorized into 4 task categories and on 13 datasets from BIG-Bench benchmark~\citep{srivastava2022beyond}, which are diverse and challenging tasks that are not encountered during training.\footnote{We exclude \textsc{Novel Concepts} task from the original T0 evaluation setting because it is a multi-label classification task. Multiple prompts are evaluated for each evaluation dataset.} We further evaluate the models on 8 new generative tasks\footnote{The dataset details of the 8 new generative tasks are provided in Appendix \ref{appen:full_list_of_datasets}.} that were not included in the original T0 paper evaluation setting. We use a \textit{rank classification} evaluation by selecting the label option with higher log-likelihood following \citet{brown2020language, sanh2021multitask} for the classification tasks. For the generative tasks, we use the ROUGE-L score as the default metric if not stated otherwise. The details of each training and evaluation dataset are provided in Appendix \ref{appen:full_list_of_datasets}.

During inference, we set $Q=32$ for applying our Retrieval-of-Expert (RoE) mechanism. We do not separately perform ablations of $S$ and $Q$, simply following the optimal setting of \citet{ye2022retrieval}.

\section{Expert LMs Can Generalize to Unseen Tasks}
\label{sec:experimental_results}
In this section, we show experimental results of expert LMs and show their potential for becoming a new paradigm over instruction tuning. Since this is a fairly novel approach of endowing LMs the capability to generalize to unseen tasks, we focus on providing \textit{proof-of-concept} of some core research questions instead of making head-to-head comparisons with all of the baselines. We leave other extensive comparisons and exhaustive ablations for future work. 

\paragraph{Main Results}
Table \ref{table:main_unseen} shows the evaluation results on the 11 unseen datasets, Table \ref{table:big_bench} shows the results on the 13 unseen BIG-Bench tasks, and Table \ref{table:generative} shows the results on the 8 unseen generative tasks. Results from the three tables show that (1) a single PE significantly outperforms T0-3B, (2) the \textsc{RoE (Orc.)} outperforms other baselines by a non-trivial margin, and (3) our simple \textsc{RoE} approach outperforms T0-3B on the classification tasks, but not on generative tasks. Details of each finding are provided in the following paragraphs. 

\textbf{\#1.} In Table \ref{table:main_unseen}, surprisingly, \textsc{T5(3B) + Cos PE}, which is a Prompt Expert (PE) that is only trained on a single prompt (`no\_prompt\_text' prompt of \textsc{Cosmos-qa} dataset), outperforms its MT LM counterpart (T0-3B) on 8 out of 11 evaluation datasets and +3.20\% on mean accuracy. Prior work shows that scaling the total number of training tasks during instruction tuning leads to better generalization; in our case, training an expert on a single task outperforms an LM trained on 300+ tasks (T0-3B). This finding is bolstered in Table \ref{table:big_bench} where the same \textsc{Cos PE} that shows the highest mean accuracy for the 11 unseen tasks outperforms T0-3B by +1.29\% on the mean accuracy performance on 13 datasets of BIG-Bench Benchmark and in Table \ref{table:generative} where \textsc{T5(3B) + Sam PE}, which is a PE trained on (`Given the above dialogue write a summary' prompt of \textsc{Samsum} dataset), outperforms T0-3B by +6.83 mean score on the 8 generative tasks. 

\textbf{\#2.} In Table \ref{table:main_unseen}, we can see that \textsc{T5(3B) + PE w/ RoE (Orc.)}, which is the upper-bound performance of choosing the best-performing expert based on the accuracy for each unseen task, outperforms T0-3B, much larger GPT-3(175B) and T0-11B by +11.94\%, +4.37\% and +2.61\%, respectively, on the mean accuracy. \textsc{T5(3B) + PE w/ RoE (Orc.)} also outperforms T0-3B by +13.69 mean score on the 8 unseen generative tasks shown in Table \ref{table:generative}. This means that \textsc{RoE} has a potential for strong unseen task generalization when the proper expert is chosen. 

\textbf{\#3.} \textsc{T5(3B) + PE w/ RoE}, which is a simple method of retrieving an expert for each unseen task leveraging an off-the-shelf retriever (Sentence Transformer~\citep{reimers-2019-sentence-bert}), outperforms T0-3B on 8 out of 11 evaluation datasets and by +2.05\% on mean accuracy. However, \textsc{T5(3B) + PE w/ RoE} underperforms T0-3B by -5.37 mean score on the 8 unseen generative tasks (Table \ref{table:generative}). Considering that \textsc{T5(3B) + PE w/ RoE} still shows a significant performance gap compared to retrieving the best-performing expert (\textsc{T5(3B) + PE w/ RoE (Orc.)}), there is much room for improvement on the retriever side. One way to close the gap is to train a \textit{supervised} retrieval model, which we leave for future work. 

\paragraph{Merging of Experts}
Table \ref{table:main_merge} shows the merging capability of expert LMs. The first three rows show the merging results of PE which are represented in the form of adapters. While \textsc{Cos\&Soc PE (Mer.)}, which is an expert constructed by performing uniform merging with \textsc{Cos PE} and \textsc{Soc PE}\footnote{\textsc{Soc PE} is a PE that was trained on \textsc{Social\_i\_qa} with prompt `no\_prompt\_text' that showed the second highest mean accuracy on the 11 unseen tasks other than PE trained with \textsc{Cosmos-qa}.} shows positive task transfers for some evaluation datasets (Copa \& Story Cloze), not all of the results are the best or second best (RTE, Hellaswag, \& Winogrande). This means that there was a negative task transfer when merging the adapter experts. 

Thus, in order to further explore the merging capability of expert LMs, we train DE via full LM fine-tuning, known to be effective in previous literature~\citep{ilharco2022editing}, and merge them as shown in the last three rows in Table \ref{table:main_merge}. \textsc{Cos De} (\textsc{Cosmos-qa}) and \textsc{Soc De} (\textsc{Social-i-qa}) are the two highest performing DE based on the mean accuracy performance on the 11 unseen tasks. While \textsc{Cos\&Soc DE (Mer.)} shows only a +0.20\% enhancement compared to \textsc{Cos DE} on mean accuracy, it still shows either the best or second best performance compared to the individual \textsc{Cos} and \textsc{Soc} DE. This implies that merging the two experts results in a composition of abilities. This opens up new possibilities of leveraging the merging of experts to unlock new capabilities which are further explored in Section \ref{sec:experiment_part2} with the composition of instructions. 

Overall, Table \ref{table:main_merge} shows that merging with adapters does not always result in positive task transfer while merging with full parameters seems to. Thus, future work should explore developing more parameter-efficient methods of merging expert LMs since always training and utilizing the entire LM weights is computationally demanding. 

\paragraph{Analysis of Experts}
\begin{table*}[ht!]
\fontsize{12}{14}\selectfont
\resizebox{\textwidth}{!}{\begin{tabular}{lcccccccc|c}
\toprule
\multicolumn{1}{c}{\multirow{2}{*}{\textbf{Method}}} & \textbf{MCQA (12)} & \textbf{Senti. (5)} & \textbf{Topic C. (3)} & \textbf{Paraph. (3)} & \textbf{STS (2)} & \textbf{Summ. (5)} & \textbf{EQA (4)} & \textbf{CBQA (2)} & \multicolumn{1}{c}{\multirow{2}{*}{\textbf{Total Avg.}}} \\ 
& (ACC) & (ACC) & (ACC) & (ACC) & (ROUGE-L) & (ROUGE-L) & (ROUGE-L) & (ROUGE-L) & \\
 \midrule
\textsc{T0-3B} & 46.97 & \underline{66.40} & 59.99 & \textbf{76.63} & 41.90 & \underline{33.10} & 28.79 & 24.67 & 47.30\\
\textsc{T0-11B} & \underline{51.32} & 64.03 & \underline{60.95} & \underline{73.64} & \underline{45.42} & \underline{33.10} & \textbf{41.20} & \underline{30.37} & \underline{50.00} \\
\textsc{T5(3B)+ PE w/ RoE} & \textbf{58.95} & \textbf{70.18} & \textbf{96.52} & 72.97 & \textbf{47.57} & \textbf{33.14} & \underline{30.36} & \textbf{51.89} & \textbf{57.70}\\
\midrule
\textsc{T5(3B)+ PE w/ RoE (Orc.)} & 56.28 & 84.52 & 96.91 & 79.34 & 47.94 & 35.40 & 40.34 & 43.24 & 60.50\\
\bottomrule        
\end{tabular}}
\caption{\footnotesize Evaluation performance on 300 sample instances from each validation dataset of the 36 training tasks categorized into 8 task categories. The number in the () represents the \# of datasets in the task category. The best comparable performances are \textbf{bolded} and second best \underline{underlined}.} 
\label{table:main_seen}
\vspace{-3mm}
\end{table*}  

\begin{table}[ht!]
\fontsize{7}{8}\selectfont
\centering
\begin{tabular}{l|cc|c}
\toprule
\multicolumn{1}{c|}{\multirow{2}{*}{\textbf{Method}}} & \textbf{Seen} & \textbf{Unseen} & \textbf{Gen} \\ 
& \textbf{Avg.} & \textbf{Avg.} & \textbf{Avg.}\\
\midrule
\multicolumn{3}{l}{\multirow{1}{*}{\textit{Before Continual Learning}}} & \textit{Unseen}\\
\midrule
 \textsc{T0-3B} & 47.30 & 51.43 & \textbf{37.98}\\
\textsc{T5(3B) + PE w/ RoE} & \textbf{57.70} & \textbf{53.48} & 32.61 \\
\midrule 
\multicolumn{3}{l}{\multirow{1}{*}{\textit{After Continual Learning}}} & \textit{Seen}\\
\midrule
\textsc{CT0-3B} & 47.54 & 50.84 & 54.52 ($\uparrow$) \\
\textsc{T5(3B) + PE$^{+}$ w/ RoE} & \textbf{57.70} & \textbf{53.33} & \textbf{55.60} ($\uparrow$)\\
\bottomrule        
\end{tabular}
\caption{\footnotesize \textbf{Seen Avg.} represents the mean accuracy of the 36 seen tasks in Table \ref{table:main_seen}. \textbf{Unseen Avg.} represents the mean accuracy of the 11 unseen tasks in Table \ref{table:main_unseen}. \textbf{Gen Avg.} represents the mean score of the 8 (unseen) generative tasks in Table \ref{table:generative}. (BS) represents BertScore. PE$^{+}$ represents augmenting the Expert Library with 8 PE trained on the 8 generative tasks. We use the LM checkpoint from \citet{chakrabarty2022finetuned} for \textsc{CT0-3B}, T0-3B continually fine-tuned on the 8 generative tasks is a sequential manner while rehearsing previous tasks. The best comparable performances are \textbf{bolded}.} 
\label{table:continual}
\vspace{-3mm}
\end{table} 

Figure \ref{fig:analysis} shows the mean accuracies of all the PE and DE results on the 11 unseen datasets. We highlight three main analyses from the figure and from the tables. 

\textbf{First}, among the 8 training task categories, Multiple-Choice Question Answering (MCQA) training tasks generally show the strongest generalization capability. We hypothesize this to be the case because all of the 11 evaluation datasets are classification tasks and require some form of question answering via instructions. This extends the findings of \citet{khashabi2020unifiedqa} that Multiple-Choice Question Answering (MCQA) generalizes well to not only different format QA tasks, but also different types of tasks such as natural language inference, story completion, coreference resolution, and word sense disambiguation as well.

\textbf{Second}, among the 36 training datasets, 3 datasets consistently ensure high performance for both PE and DE: \textsc{Cosmos-qa}~\citep{huang-etal-2019-cosmos}, \textsc{Social-i-qa}~\citep{sap-etal-2019-social}, and \textsc{Dream}~\citep{sun-etal-2019-dream}. All three datasets are commonsense reasoning datasets, which have been considered to be crucial for generalization to unseen tasks~\citep{lourie2021unicorn}. We provide the full ranking of the PE and DE for the 11 unseen tasks shown in Figure \ref{fig:analysis} in Appendix \ref{appen:expert_rankings}.

\textbf{Lastly}, \textsc{T5(3B) + Sam PE} which is a PE trained on \textsc{Samsum}~\citep{gliwa-etal-2019-samsum}, a dataset with abstractive dialogue summaries, shows the best mean score on the 8 unseen generative tasks in Table \ref{table:generative}, outperforming T0-3B by +6.83 mean score. However, the same PE shows one of the lowest ranks for the 11 unseen (classification) tasks (shown in Appendix \ref{appen:expert_rankings}) underperforming T0-3B by -9.15\% mean accuracy. This shows that there is \textit{no free lunch}: a PE that shows high mean performance for unseen generative tasks do not show high mean performance for unseen classification tasks. This also implies that it is more-so important to retrieve the correct expert dynamically depending on the given context (target task). 


\section{Benefits of Expert LMs over MT LMs} 
\label{sec:experiment_part2}
In this section, we highlight the 3 main benefits of expert LMs and \textsc{RoE} over MT LMs.

\paragraph{Seen Task Performance}
First, we show that expert LMs are less susceptible to negative task transfer by comparing the performance of \textsc{T5(3B) + PE w/ RoE} on the validation datasets of the 36 training datasets with two MT LMs, T0-3B and T0-11B. As shown in Table \ref{table:main_seen}, our distributed approach outperforms T0-3B and T0-11B by +10.40\% and +7.70\% on mean accuracy, respectively. 

This is because since evaluation is done with \textit{seen} instructions, our simple retrieval mechanism is highly likely to select the best-performing expert from the Expert Library, showing comparable performance to \textsc{T5(3B) + PE w/ RoE (Orc.)}. In fact, \textsc{T5(3B) + PE w/ RoE} retrieves the PE from the same training dataset on 280 out of 296 seen tasks, and the PE trained with both the same prompt and dataset (oracle) on 185 out of 296 seen tasks. 


\paragraph{Continual Learning of New Tasks}
In some scenarios when we want to additionally fine-tune LMs on additional datasets \textit{after} model deployment, making finetuned LMs continual learners is important~\citep{chakrabarty2022finetuned}. This is because performing instruction tuning on the entire set of original and additional tasks in each update would lead to heavy computation. Previous work mitigates this issue through a rehearsal-based method, continually training the instruction-tuned LM on \textit{samples} of the original and additional tasks~\citep{chakrabarty2022finetuned}. However, this approach (1) assumes that we have access to the original datasets and (2) still leads to additional computational overhead, especially when scaling the total number of seen tasks during instruction tuning. 

We show that we can accomplish the same feat through distributed training of experts without any access to original, seen datasets by training separate experts for each additional task and simply adding them to the Expert Library. Specifically, we show the comparison between continually training an MT LM (T0-3B) which is referred to as \textsc{CT0-3B} through a rehearsal-based approach, and our distributed approach on 8 new generative tasks in Table \ref{table:continual}. The 8 generative tasks for continual learning were chosen following the previous work~\citep{chakrabarty2022finetuned}. 


The table shows that our distributed approach results in absolutely no degradation of performance for the seen task, a minor (-0.15\%) degradation for unseen tasks, and superior mean performance (+1.08) for the 8 target tasks compared to the MT LM counterpart, outperforming \textsc{CT0-3B} on 7 out of the 8 target tasks. This shows that without any access to original datasets or heavy computational cost, our distributed approach is mostly able to retain its original ability (seen \& unseen) as well as outperform CT0-3B on the target tasks. 
We leave scaling the number of new target tasks and how our distributed approach performs against its instruction-tuned counterpart for future work.

\paragraph{Compositional Instructions}
\begin{table}[t!]
\fontsize{12}{14}\selectfont
\resizebox{\columnwidth}{!}{\begin{tabular}{l|ccccc|c}
\toprule
\multicolumn{1}{c|}{\multirow{2}{*}{\textbf{Method}}} & \textbf{xsum} & \textbf{xsum} & \textbf{xsum} & \textbf{xsum} & \textbf{xsum} & \textbf{Total}\\
& \textbf{en$\rightarrow$ko} & \textbf{en$\rightarrow$es} & \textbf{en$\rightarrow$zh} & \textbf{en$\rightarrow$fr} & \textbf{en$\rightarrow$ja} & \textbf{Avg.}\\
\midrule
\textsc{mT0-3B} & 1.84 & 16.14 & \textbf{6.74} & 20.37 & 3.44 & 9.71\\
\textsc{mT5-3B + Mer. Ex.} & \textbf{8.23} & \textbf{16.97} & 2.40 & \textbf{20.55} & \textbf{13.98} & \textbf{12.43}\\
\bottomrule        
\end{tabular}}
\caption{\footnotesize Comparison of compositional abilities of both summarization and translation task for MT LM (\textsc{mT0-3B}) and our distributed approach (\textsc{mT5-3B + Mer. Ex.}) which involves merging the corresponding experts. ROUGE-L is used as the evaluation metric. ko, es, zh, fr, ja stand for Korean, Spanish, Chinese, French, and Japanese, respectively. The best comparable performances are \textbf{bolded}} 
\label{table:composition}
\end{table} 

\begin{table}[t!]
\fontsize{12}{14}\selectfont
\resizebox{\columnwidth}{!}{
\begin{tabular}{c|l}
\toprule
\textbf{Task} & \textbf{\textsc{example}} \\
\midrule
\textbf{xsum} & \textbf{\textsc{mT0}}:  El asesinato de un niño de tres años de edad en Francia fue atribuido a su hermano \\
\textbf{en$\rightarrow$es} & mayor. \\
 &\textbf{M.E.}: La policía francesa arrestó a cuatro miembros de la familia del niño por su presunta \\
& implicación en el crimen, que ocurrió el 17 de septiembre en la casa familiar en Mulhouse, \\
& al oeste del país, y de más de 100.000 habitantes. \\
& \textbf{G.T.}: La polic ́ıa que investiga el fallecimiento de un ni  ̃no de 9 a  ̃nos en Francia,\\
& supuestamente golpeado hasta la muerte, arrest  ́o este martes a cuatro miembros de su \\
& familia, seg  ́un declaraciones de los fiscales a la agencia de noticias AFP.\\
\midrule

\textbf{xsum} & \textbf{\textsc{mT0}}:  Le président de la République démocratique du Malawi a été condamné à cinq ans \\
\textbf{en$\rightarrow$fr} & de prison pour complicité dans l'assassinat de Paul Mphwiyo. \\
&\textbf{M.E.}: Le 8 novembre 2013, l'ancien ministre de la Justice du Malawi, M. Ralph\\
& Kasambara, a été arrêté après avoir commis le meurtre de Paul Paul MPHWIYO, le \\
& directeur du budget du ministère des Finances. \\
& \textbf{G.T.}: La Haute Cour de Lilongwe a condamn  ́e mardi l’ancien ministre de la Justice, \\
& Raphael Kasambara, `a 13 ans d’emprisonnement et de travaux forc  ́es pour complicit \\
&́e de meurtre \\
\midrule

\textbf{xsum} & \textbf{\textsc{mT0}}:  华为副总裁 Meng Ship 和副总统 Meng Teng 被加拿大警方逮捕,并被指控侵犯公民权利。\\
\textbf{en$\rightarrow$ja} &\textbf{M.E.}: カナダの最高裁判所(CFO)は、12月に逮捕された創設者の息子であり、\\
&副社長はカナダ政府とカナダ移民局(CBSA)と警察を告訴した。  \\
& \textbf{G.T.}: 中の通信機器最大手、華技術(ファウェイ)の最高財務責任者(CFO)の孟晩\\
& 舟副長は、昨年12月にカナダ局がアメリカの要請で自分を逮捕したことについて、\\
& カナダを提訴した。\\
\midrule

\textbf{xsum} & \textbf{\textsc{mT0}}:  The Sierra Leonean nurse who was isolated for seven hours at the airport terminal\\
\textbf{en$\rightarrow$zh} & has said that the isolation experience is "terrifying" and may make other medical workers\\
&  reluctant to go to West Africa.\\
&\textbf{M.E.}: 一名感染埃博拉病毒的医生Craig Spencer目前正在纽约大都会医院接受隔离治疗,但她只得到了\\
& 一个粮食棒来填补她的饥饿。 \\
& \textbf{G.T.}: 一位曾在塞拉利治埃博拉病人的美士返回美后被隔察,批了瓦克机待的方式\\
\midrule

xsum & \textbf{\textsc{mT0}}: Korean peninsula has had its warmest winter since 1973, according to the \\
\textbf{en$\rightarrow$ko} & Meteorological Administration. \\
&\textbf{M.E.}: 지난해 1월은 국내에서 가장 따뜻한 겨울이었다.\\
& \textbf{G.T.}: 올겨울, 추위가 실종됐다. 따뜻한 날씨가 이어지면서 눈 구경도 어려워졌다.\\

\bottomrule
\end{tabular}
}
\caption{\footnotesize Example outputs from the 5 Compositional Tasks given the input ``\textit{Write a summary of the following English text and translate the sentence into [Language]: [English Summary]}.". \textbf{M.E.} stands for Merged Experts. \textbf{G.T.} stands for Ground Truth. es, fr, ja, zh, and ko stand for Spanish, French, Japanese, Chinese, and Korean, respectively. The actual input for the examples are provided in Appendix \ref{appen:composition_details}.} 
\label{table:composition_examples}
\end{table} 

Prior work has shown the need for performing \textit{compositional} instructions~\citep{logeswaran2021learning, corona-etal-2021-modular, khot2022decomposed}. For example, we can give the following instruction to the LM: ``\textit{Write a summary of the following English text and translate the sentence into Korean.}" where ``\textit{Write a summary of the following English text.}" and ``\textit{Translate the sentence into Korean.}" are two separate instructions seen during training. To test this compositional capability, especially in a multi-lingual setting, we utilize the mT0-3B~\citep{muennighoff2022crosslingual} as our MT LM and evaluate the composition of performing 5 novel compositional tasks of summarization and translation. To explore the benefits of merging experts for performing compositional instructions, we perform 6 full fine-tuning with mT5-3B~\citep{xue-etal-2021-mt5} as the underlying vanilla pretrained multilingual LLM: We use \textsc{xsum} to train one English Summarization expert and use five translation pairs in \textsc{tatoeba} (en$\rightarrow$es, en$\rightarrow$fr, en$\rightarrow$ja, en$\rightarrow$zh, en$\rightarrow$ko) to train the corresponding five translation experts. During inference, we merge the summarization expert with each of the five translation experts\footnote{We provide the specific configurations used for merging such as the $\lambda_{i}$ values for each task vector $\tau_{i}$ and the training and validation stats in Appendix \ref{appen:composition_details}}. Note that both \textsc{xsum} and \textsc{tatoeba} are part of the training tasks used during instruction tuning of mT0-3B. 

Evaluation results on the five compositional tasks are shown in Table \ref{table:composition}. Our distributed approach, \textsc{mT5-3B + Mer. Ex.}, outperforms its MT LM counterpart, \textsc{mT0-3B} on 4 out of the 5 tasks and by a mean ROUGLE-L score of +2.71; This is due to a significant performance gap for the tasks involving low-resource languages (Korean and Japanese) because the low-resource languages are protected from negative transfer when doing distributed training. 
Cherry-picked output examples of the MT LM and the merged experts are provided in Table \ref{table:composition_examples}.

\section{Limitations and Discussions}
\label{discussion_and_Limitation}
While we highlight some of the major drawbacks of instruction tuning and propose an alternative approach of instead training and retrieving experts in this paper, we do not perform experimental results over MT LMS that have more than $>$11B parameters. For example, MT LMs with $>$11B parameters may be less susceptible to negative task transfer because of increased model capacity. Also, during the inference of unseen tasks, our retrieval mechanism assumes batch inference (i.e. having access to 32 samples of the target tasks without labels). Finally, when showing the compositional instruction experiments, we assume the two optimal experts could be retrieved from the compositional instruction (concatenation of the two seen instructions) given as the input along with the evaluation instance. This might not necessarily be the case with more complex, compositional instructions, which might require a separate \textit{decomposition} stage. We instead focus on showing the possibility merging experts can bring and leave developing novel methods of retrieving the optimal experts during inference for future work. 

\section{Conclusion}
\label{conclusion}
In this work, we provide an interesting finding that \textit{expert} LMs trained on single tasks show strong generalization capability to unseen tasks, even surpassing MT LMs trained on multiple tasks (300+) by a non-trivial margin. We leverage this capability and show three main benefits of training and retrieving experts for inference over MT LMs, demonstrating that our proposed distributed approach is more robust against negative task transfer, more adapt at learning new tasks, and can perform compositional instructions. To this end, we urge the research community to further explore distributed and collaborative training of experts which may have other future benefits including efficiency, privacy, and personalization not explicitly explored in this paper. 

\subsubsection*{Acknowledgments}
We thank Colin Raffel, Sungdong Kim, Sejune Joo, Miyoung Ko, Eunbi Choi, Hyunji Lee, Dongkeun Yoon, Yoonjoo Lee, and Yujin Kim for the useful discussion and feedback on the paper. 


\bibliography{example_paper}
\bibliographystyle{icml2023}

\newpage
\appendix
\onecolumn
\section{Details of Training and Evaluation Datasets}
\label{appen:full_list_of_datasets}
\paragraph{Details of Training Dataset}
Following \citet{sanh2021multitask}, we use 36 training datasets from the 8 task categories for training our experts. We provide the official names given in Huggingface Datasets: \textbf{Sentiment Classification (Senti.)}
imdb \cite{maas-etal-2011-learning}, amazon\_polarity \cite{McAuley2013HiddenFA}, rotten\_tomatoes \cite{pang-lee-2005-seeing}, yelp\_review\_full \cite{zhang2015character}, and app\_reviews. \textbf{Paraphrase Identification (Para.)}
glue/qqp \cite{2018arXiv180407461W}, glue/mrpc \cite{2018arXiv180407461W}, and paws/labeled\_final \cite{zhang-etal-2019-paws}. \textbf{Topic Classification (Topic C.}
ag\_news \cite{NIPS2015_250cf8b5}, dbpedia\_14 \cite{Lehmann2015DBpediaA}, and trec \cite{li-roth-2002-learning}. \textbf{Summarization (Summ.)}
gigaword \cite{graff2003english}, multi\_news\ \cite{fabbri-etal-2019-multi}, samsum \cite{gliwa-etal-2019-samsum}, xsum \cite{narayan-etal-2018-dont}, and cnn\_dailymail/3.0.0 \cite{see-etal-2017-get}. \textbf{Structure-To-Text (STS)}
common\_gen \cite{lin-etal-2020-commongen} and wiki\_bio \cite{lebret-etal-2016-neural}. \textbf{Multiple-Choice Question Answering (MCQA)}
commonsense\_qa \cite{talmor-etal-2019-commonsenseqa}, dream \cite{sun-etal-2019-dream}, quail \cite{DBLP:conf/aaai/RogersKDR20}, qasc \cite{allenai:qasc}, quarel \cite{tafjord2019quarel}, 
 cos\_e/v1.11 \cite{rajani2019explain}, quail \cite{Rogers_Kovaleva_Downey_Rumshisky_2020}, social\_i\_qa \cite{sap-etal-2019-social}, wiqa \cite{tandon-etal-2019-wiqa}, cosmos\_qa \cite{huang-etal-2019-cosmos}, sciq \cite{welbl-etal-2017-crowdsourcing}, and wiki\_hop/original \cite{welbl-etal-2018-constructing} \textbf{Extractive Question Answering (EQA)}
adversarial\_qa/adversarial\_qa \cite{bartolo2020beat}, quoref \cite{bartolo-etal-2020-beat}, ropes \cite{lin-etal-2019-reasoning}, and duorc/Paraphrase IdentificationRC \cite{saha-etal-2018-duorc} 
 \textbf{Closed Book Question Answering (CBQA)}
kilt\_tasks/hotpotqa \cite{kilt_tasks} and wiki\_qa \cite{yang-etal-2015-wikiqa}.

\paragraph{Details of Evaluation Dataset}
Following \citet{sanh2021multitask}, we include 11 evaluation datasets as follows: RTE \cite{dagan2005pascal}, CB \cite{de2019commitmentbank}, ANLI \cite{nie2019adversarial} for natural language inference task, COPA \cite{roemmele2011choice}, Hellaswag \cite{zellers2019hellaswag}, Storycloze \cite{mostafazadeh2016corpus} for sentence completion task, Winogrande \cite{sakaguchi2021winogrande}, WSC \cite{levesque2012winograd} for coreference resolution task, and WiC \cite{pilehvar2018wic} for word sense disambiguation task.

For BIG-bench tasks, we evaluate on 13 tasks, following \citet{sanh2021multitask}: Known Unknown, Logic Grid, StrategyQA, Hindu Knowledge, Movie Dialog, Code Description, Conceptual, Language ID, Vitamin C, Syllogisms, Misconceptions, Logical Deduction, and Winowhy.

For the generative evaluation tasks, we follow \citet{chakrabarty2022finetuned} and utilize 8 tasks: Text Simplification (WikiAuto)~\citep{jiang-etal-2020-neural}, Headline Generation with constraint (HGen)~\citep{yamada-etal-2021-transformer}, Haiku Generation (Haiku), Covid QA~\citep{moller-etal-2020-covid}, Inquisitive Question Generation (ELI5)~\citep{fan-etal-2019-eli5}, Empathetic Dialogue Generation (EmDg)~\citep{rashkin-etal-2019-towards}, Explanation Generation (eSNLI)~\citep{NIPS2018_8163}, and Twitter Stylometry (Twitter)

\section{Varying the Embedding Model and Text Format for Retrieval of Experts}
\label{appen:text_format}

\paragraph{Performance of Different Embedding Models} While \citet{ye2022retrieval} used T0~\citep{sanh2021multitask} as the base embedding model to retrieve prompt embeddings, we explore 13 different sentence embedding models to waive the need of using instruction tuned models for retrieval of expert LMs. 

\begin{table*}[t!]
\fontsize{12}{14}\selectfont
\resizebox{\textwidth}{!}{\begin{tabular}{lccccccccccc|c}
    \toprule
     \textbf{Embedding Models} & \textbf{Hellasw.} & \textbf{StoryC.} & \textbf{AN. R1} & \textbf{AN. R2} & \textbf{AN. R3} & \textbf{COPA} & \textbf{CB} & \textbf{RTE} & \textbf{WSC} & \textbf{WiC} & \textbf{Winogr.} & \textbf{Total Avg.}\\ 
    \midrule
    \textsc{Random} & 31.25	&47.38	&32.94	&33.38&	32.12&	61.00	&38.57&	54.01&	46.35	&49.03&	54.27&43.66\\
    \midrule
    \textsc{all-MiniLM-L6-v2} &\underline{34.60}	&\underline{86.33}	&\textbf{35.49}	&\textbf{34.64}	&31.22	&79.25&	43.57	&\textbf{64.01}	&\underline{62.21}&	\underline{52.97}&	\textbf{61.60}&\textbf{53.48}\\
    \textsc{all-MiniLM-L12-v2} &32.33&	67.13&	33.84&	33.38&	33.69&	63.00&	\underline{47.38}&	58.48&	49.52&	51.17&	\underline{56.80}&47.88\\
    \textsc{all-mpnet-base-v2} &31.53&	59.33&	33.71&	33.02	&31.73&	61.38&	46.43&	53.97&	44.62&	52.33&	54.93 & 47.73 \\
    \textsc{nli-mpnet-base-v2} &22.60&	50.87&	34.02	&33.69	&\underline{34.53}&	58.75&	38.57&	48.59&	52.21&	49.77&	51.07 & 43.15\\
    \textsc{sup-simcse-roberta-large} &26.93	&59.67&	34.58&	33.29&	\textbf{34.73}&	84.75&	41.90	&52.06&	50.67&	\textbf{56.03}&	51.67&47.84\\
    \textsc{unsup-simcse-roberta-large} &24.27	&71.93&	33.98&	32.22&	33.78&	69.75&	43.33&	50.72&	55.38&	50.33&	50.93&46.97\\
    \textsc{hkunlp/instructor-large} &19.80	&57.33&	33.16	&\underline{33.78}	&32.93	&54.50	&39.64	&47.80	&55.96	&49.20	&51.20&43.21\\
    \textsc{hkunlp/instructor-xl} &19.60	&44.53	&32.62	&32.82	&32.31	&57.88	&44.52	&47.83	&60.77	&48.77	&51.80&43.04\\
    \textsc{gtr-t5-large} &29.60	&70.47	&33.04	&31.64	&32.31	&58.38	&\textbf{50.95}	&54.69	&57.79	&51.50	&50.80&47.38\\
    \textsc{gtr-t5-xl} &\textbf{37.20}	&84.80	&33.24	&33.27	&33.58	&\underline{83.00}	&43.69	&58.59	&45.00	&50.73	&51.07&50.38\\
    \textsc{sentence-t5-large}&33.33	&78.53	&33.11	&33.76	&33.31	&\textbf{87.25}	&46.19	&58.34	&\textbf{63.08}	&52.13	&54.27&\underline{52.12} \\
    \textsc{sentence-t5-xl} &25.67	&\textbf{87.13}	&\underline{35.27}	&33.38	&32.98	&68.63	&46.19	&\underline{59.10}	&61.63	&52.33	&51.67&50.36\\
    \textsc{voidism/diffcse-bert-base-uncased-sts} &21.93	&46.53	&33.07	&32.91	&32.47	&58.75	&45.60	&49.71	&60.77	&49.70	&50.33&43.80\\
    \midrule
    \textsc{T0-small~\citep{ye2022retrieval}}&39.55	&97.09	&33.89	&33.96	&34.38	&88.00&	41.55	&62.53	&53.95&52.45	&70.20&55.23\\
    \bottomrule        
\end{tabular}}
\caption{\footnotesize Comparison of different embedding models, measured on 11 different unseen datasets using Prompt Experts(PE). For instance, \textsc{all-MiniLM-L6-v2} refers to \textsc{T5(3B) + PE w/ RoE} in Table~\ref{table:main_unseen}. All the task format are fixed to \textit{`Answer Choices: \{answer choice\}, Instance: \{instance\}'}. The best comparable performances are \textbf{bolded} and second best \underline{underlined}. Note that evaluation is performed with 300 samples from each evaluation dataset for efficiency.}
\label{table:embedding_ablation}
\vspace{-3mm}
\end{table*}  

More specifically, we list of embedding models we use are as follows: (a) 4 different variants of \textsc{Sentence Transformer} model~\citep{reimers-2019-sentence-bert}: all-MiniLM-L6-v2, all-MiniLM-L12-v2, all-mpnet-base-v2, nli-mpnet-base-v2, (b) 2 different variants of \textsc{SimCSE} model~\citep{gao2021simcse}: sup-simcse-roberta-large, unsup-simcse-roberta-large, (c) 2 different variants of \textsc{Instructor} model~\citep{su2022one}: hkunlp/instructor-large, hkunlp/instructor-xl, (d) 2 different variants of \textsc{GTR} model~\citep{ni2021large}: gtr-t5-large, gtr-tr-xl, (e) 2 different variants of \textsc{SentenceT5} model~\citep{ni2022sentence}: sentence-t5-large, sentence-t5-xl, and (f) \textsc{DiffCSE} model~\citep{chuang2022diffcse}: voidism/diffcse-bert-base-uncased-sts which are all available on HuggingFace. Note that we try different embedding models in an unsupervised manner, i.e., not requiring any supervision to train the embedding model, but using it off-the-shelf. The results are shown in Table~\ref{table:embedding_ablation}.

\paragraph{Performance of Different Text Formats} We also try different variants of text format given to the embedding model. Using Promptsource~\citep{bach2022promptsource}, we compare including the instance, label list, answer choice in 2 different formats. Specifically, the full list of text formats are as follows: (a) \textit{`Instance: \{instance\}'}, (b) \textit{`Answer Choices: \{label list\}'}, (c) \textit{`Answer Choices: \{answer choice\}'}, (d) \textit{`Answer Choices: \{label list\}, Instance: \{instance\}'}, (e) \textit{`Answer Choices: \{answer choice\}, Instance: \{instance\}'}, (f) \textit{`\{instance\}'}, (g) \textit{`\{label list\}'}, (h) \textit{`\{answer choice\}'}, (i) \textit{`\{label list\}$<$/s$>$\{instance\}'}, (j) \textit{`\{answer choice\}$<$/s$>$\{instance\}'}. Label list and answer choice differ in that while label list uses the actual label options (e.g., [`swim',`fly',`walk',`run']), answer choice organizes them with a `|' deliminator in the middle (e.g. A$|$B$|$C$|$D). The results are shown in Table~\ref{table:textformat_ablation}.

\paragraph{Results} While we tried different variants, the oldest, yet most chosen model \textsc{all-minilm-l6-v2} outperforms other options. We conjecture that this is because most of the model variants we tested were trained as sentence embedding models, not for embedding prompted instances. Prompted instances are some how structural and formatted compared to natural language sentences used for training sentence embedding models. In terms of text format, using both the prompted instance and the answer choice showed the best results. These results show that for the dense retriever to map instances, it should rely on both components, which are orthogonally important. Also, using the actual label option harms performance compared to using the answer choice, which indicates that the output format itself is important to retrieve well-matched expert LMs.

\begin{table*}[t!]
\fontsize{12}{14}\selectfont
\resizebox{\textwidth}{!}{\begin{tabular}{lccccccccccc|c}
    \toprule
     \textbf{Text Format} & \textbf{Hellasw.} & \textbf{StoryC.} & \textbf{AN. R1} & \textbf{AN. R2} & \textbf{AN. R3} & \textbf{COPA} & \textbf{CB} & \textbf{RTE} & \textbf{WSC} & \textbf{WiC} & \textbf{Winogr.} & \textbf{Total Avg.}\\ 
    \midrule
    \textit{`Instance: \{instance\}'} &24.67&	78.07&	33.53&	32.67&	32.91	&64.13&	40.36	&54.55	&50.48	&52.47	&52.73&46.96\\
    \textit{`Answer Choices: \{label list\}'} &24.93	&51.47	&33.80&	34.29	&33.20	&58.38&	42.38	&50.83	&51.54	&\underline{53.47}	&51.13&44.13\\
    \textit{`Answer Choices: \{answer choice\}'} &31.60&	50.53	&32.09	&32.16	&\textbf{35.98}&	\underline{84.75}	&44.05	&50.83	&51.54	&\underline{53.47}	&\textbf{63.40}&48.22\\
    \textit{`Answer Choices: \{label list\}, Instance: \{instance\}'} &32.27&	56.40	&\textbf{35.76}	&\textbf{34.73}&	31.11	&67.13	&\underline{46.31}	&59.17&	61.15	&52.30&	52.67&48.09\\
    \textit{`Answer Choices: \{answer choice\}, Instance: \{instance\}'} &\underline{34.60}	&\textbf{86.33}	&\underline{35.49}	&\underline{34.64}	&31.22	&79.25&	43.57	&\textbf{64.01}	&62.21&	52.97&	\underline{61.60}&\textbf{53.48}\\
    \textit{`\{instance\}'} &24.27&	\underline{82.40}	&33.53	&33.47&	\underline{33.89}	&58.25	&43.81	&51.66	&51.92	&52.60	&51.13&46.99\\
    \textit{`\{label list\}'} &24.53	&50.53	&33.67	&32.76	&32.58&	58.38	&42.02&	50.83&	51.54&	\underline{53.47}&	51.13&43.77\\
    \textit{`\{answer choice\}'} &24.00	&49.87	&32.09	&32.16&	\textbf{35.98}	&\textbf{86.00}&	44.05	&50.83	&51.54	&\underline{53.47}	&\textbf{63.40}&47.58\\
    \textit{`\{label list\}$<$/s$>$\{instance\}'} &25.53&	65.60	&\textbf{35.76}	&33.91	&31.07&	62.38	&\textbf{46.90}	&\underline{60.14}	&
    \textbf{62.69}	&\textbf{53.70}	&50.73&48.04\\
    \textit{`\{answer choice\}$<$/s$>$\{instance\}'} &\textbf{35.93}	&60.53	&35.29&	32.51&	33.00&	68.75&	43.93	&59.03&	\underline{62.60}	&52.40	&60.73&\underline{49.52}\\
    \bottomrule        
\end{tabular}}
\caption{\footnotesize Comparison of different text formats, measured on 11 different unseen datasets using Prompt Experts(PE). For instance, \textit{`Answer Choices: \{answer choice\}, Instance: \{instance\}'} refers to \textsc{T5(3B) + PE w/ RoE} in Table~\ref{table:main_unseen}. All the embedding model are fixed to \textsc{all-MiniLM-L6-v2}. The best comparable performances are \textbf{bolded} and second best \underline{underlined}. Note that evaluation is performed with 300 samples from each evaluation dataset for efficiency.}
\label{table:textformat_ablation}
\vspace{-3mm}
\end{table*}

\section{Details of Performing Compositional Instructions}
\label{appen:composition_details}
Our compositional instruction setting consists of a total of 400 instances for each task (300 instances for the validation set, and 100 instances for the test set.) per language that was obtained using google translate to change the input of the \textbf{XL-Sum}~\citep{xlsum} dataset. We thus use the ground truth label in the specified language and the input is the machine-translated version. The reason for this is that we measure the $\lambda_{i}$ values (the importance to place on each task vector $\tau_{i}$) by performing evaluation on the validation datasets. Empirically, setting 1.0 for each $\lambda_{i}$ value resulted in the best performance. Thus, as mentioned in the method section, the total $\sum\lambda_{i}$ results in 2.0, greater than 1.0. 

We also vary the decoding strategies to check the performance of merging two experts finetuned from \textsc{mT5-3B} compared with naive \textsc{mT0-3B} on \textbf{XL-Sum} dataset. The detailed optimal setting we found is as follows:

\begin{itemize}
\item \textsc{lambda1}: 1.0
\item \textsc{lambda2}: 1.0
\item \textsc{no\_repeat\_ngram\_size}:2
\item \textsc{temperature}:1.0
\item \textsc{early\_stopping}:True
\item \textsc{do\_sample}:True
\item \textsc{top\_p}:0.95
\end{itemize}

Here are the actual inputs for the LM generated \& ground truth output examples shown in Table \ref{table:composition_examples}. The \textit{compositional} instruction portion is shown in \textbf{bold}.

\textsc{English $\rightarrow$ Spanish:}
``\textbf{Write a summary of the following English text and translate the sentence into Spanish:} The French police arrested four members of the child's family for their alleged involvement on Tuesday. Police sources told local media that the child refused to do his homework and that he was beaten with the stick of a broom. The 20 -year -old sister, his older brother and his girlfriend were present at the time of the incident and were arrested. The three called emergency services, which could not save the child. The alleged crime occurred on September 17 at the family's home in the town of Mulhouse, in the east of the country, and of just over 100,000 inhabitants. Although the child's mother was not at home because she was on a trip for work reasons, she was also arrested. The authorities say it will be questioned to confirm whether it encouraged the punishment. The four family members remain in police custody and must appear before the Mulhouse Prosecutor's Office for a judicial investigation. Prosecutor Edwige Roux-Morizot will investigate the case. Moretones after the death of the child, victim of cardiac arrest, several neighbors celebrated a vigil in their honor and met with the child's parents to offer them comfort. However, the results of the autopsy motivated the police to carry out an investigation into what happened. The child's body presented several bruises, especially at his feet, according to AFP. Despite the confirmation of cardiac arrest, pathologists said the cause of death was probably the blows he had suffered. A police source said the child was beaten with blunt objects. Although the main suspect of the murder is the older brother, the French authorities hope that the investigation will shed light on what happened. France is one of the 13 countries of the European Union where corporal punishment is legal. A legal practice The National Assembly of France is considering approveing a law to prohibit corporal punishments for children. There are two new law proposals that would grant children a violence -free education, venting parents to use "forms of humiliation such as physical or verbal..."

\textsc{English $\rightarrow$ French:} 
``\textbf{Write a summary of the following English text and translate the sentence into French:} The former Minister of Justice of Malawi, Ralph Kasambara, was arrested on November 8, 2013. Mr. Kasambara was found guilty of conspiracy in the assassination in September 2013 of the former budget director at the Ministry of Finance, Paul Paul MPHWIYO. The murder of Mr. Mphwiyo had led to the discovery of the scandal of "cashgate", the systematic looting of public resources, during the administration of President Joyce Banda. Nearly 250 million had been fraudulently paid to businessmen for services who have never been rendered. A few days before the tragedy, a subordinate official would have been found with gold bars belonging to the cash, the equivalent of more than \$ 300 million, in the trunk of his car. Money was also confiscated at the home of certain officials and in chests from their vehicles. Immediately after his conviction last month, Kasambara had suggested that he would not appeal the court verdict."

\textsc{English $\rightarrow$ Japanese:} 
``\textbf{Write a summary of the following English text and translate the sentence into Japanese:} Vice Chairman Meng Ship, the highest financial manager (CFO), was the daughter of the founder arrested in Vancouver, Canada last December, and Vice President Meng Teng was sanctioned at Vancouver Airport last December. He was arrested for violating and associated scams and was charged at the end of January this year. The United States authorities are seeking to hand over the vice chairman, but they deny the charges. Defendant Meng filed an administrative lawsuit for the Canadian government, the immigration bureau, and the police for "significantly infringing" their citizenship. China has accused the defendant's arrest and delivery procedure as a "political project." <Related article> Introduction is "illegal" and "Dandridy" British Columbia Senior Court on the 1st, and Meng is the Canadian government and the Royal Canadian equestrian police (RCMP), and the Canadian Immigration Bureau (CBSA). He is complaining of civil rights infringement. Before the arrest of RCMP, CBSA complained that he had detained himself on unfair claims, investigated and interrogated his belongings. The vice chairman was bail and was at Vancouver's home, and the authorities arrested Vice Chairman Meng on the spot. He complained that it infringed on the rights based on the Canadian Characters of Human Rights. In addition, Vice -Chairman's detention was "illegal" and "arbitrary", and authorities pointed out that "the reason for detention, the right to call lawyers, or the right to be paid to be silent." What is the reaction of each country? The relationship between China, Canada and the United States has deteriorated over the arrest of Vice Chairman Meng. In January, the U.S. Department was charged with 23 cases of Huawei and Vice Chairman Meng. In addition to bank fraud, communication fraud, judicial obstruction, a major US telecommunications equipment T -mobile has been charged with trying to steal technology. China accused these movements as "abuse of the handover agreement" between the United States and Canada, and stated that they..."

\textsc{Englsih $\rightarrow$ Chinese:} 
``\textbf{Write a summary of the following English text and translate the sentence into Chinese:} Dr. Craig Spencer, who is infected with Ebola virus, is currently being hospitalized at the New York Metropolitan Hospital. Caisyex said that the isolation experience is very scary and may also make other medical workers reluctant to go to West Africa to help curb the Ebola epidemic. Following New York and New Jersey, Illinois has also adopted a strict isolation policy. New measures means that those who have come into contact with any Ebola patient in West Africa will be forced to isolate for 21 days. U.S. President Obama Obama said in a weekly radio speech on September (October 25) that Americans should believe in the facts rather than being dominated. He also reiterated that he can infect the virus only with direct body fluids with Ebola patients. Higkos, who was "criminals", who was an isolated person, said that she had witnessed "confusion, panic, and the most terrifying isolation" when she returned from Sierra Leone on Friday (24th). Hekox wrote a newspaper in the United States: "I don't know how many medical workers who fought with Ebola virus in the West African epidemic area will have the same encounter." She said, "Will they feel like criminals like criminals like criminals? She also said that she was isolated for seven hours at the airport terminal, but she only got a grain rod to fill her hunger. She denied that she had had a fever and said that she was just blushing at the time because she was not satisfied with the treatment at the airport. Even though Hiccoks was negative in Ebola virus testing, she was still being isolated for three weeks and was monitored by medical officials. Frontline medical staff was deeply influenced by the Ebola outbreak. After being diagnosed with Ebola patients, a doctor of New York, who had worked in Guinea last week, was diagnosed with Ebola patients, New York State and New Jersey have strengthened their isolation measures. Spencer is currently receiving isolation treatment in a hospital in New York. Mali has also recently appeared in Ebola, and President Ibrahim... "

\textsc{English $\rightarrow$ Korean:} 
``\textbf{Write a summary of the following English text and translate the sentence into Korean:} According to the Korea Meteorological Administration, January this year was the warmest winter since 1973, when the weather observation began in the Korean peninsula. The average temperature in January last month was 2.8 degrees. This is 3.8 degrees higher than the average of minus 1.0 degrees in January, 1981 ~ 2010. The previous average temperature record was 1.6 degrees in 1979. Except for the first day of the new year, the average temperature in the country was higher than normal. Due to the high temperature, the snowfall was the lowest. The Korea Meteorological Administration cited the introduction of warm southwestern air flow into the Siberian region, and the fact that the 'pole whirl', which traps cold air in the Arctic, was strong as an abnormal temperature. It also analyzed that the warm south wind flow was introduced to the Korean peninsula due to the high sea level temperature of the Western Pacific. Nationwide weather data in January, the average temperature in the coldest January of the year has continued to rise in recent years. According to the weather data released by the Korea Meteorological Administration in January 1973-2020, the average temperature in January in Korea is steadily rising. Choi Jung -hee, the Korea Meteorological Agency, said that the warming of winter is "global warming impact," and "most of the monthly weather data tends to be similar." Detection of the ecosystem change is detected throughout the ecosystem. The first spawning season of 'Bukbangsan Guri', a climate change indicator, has been faster. Mudeungsan National Park Eastern Office said on the 24th of last month that the first spawning of the North Bangsan Gogi, a species designated by the Ministry of Environment, was observed. It was first observed. It is 27 days earlier than February 19 last year. This is the first time that spawning has been observed in January since 2010, when the survey began. Researchers at the Park Industrial Complex believed that the spawning day was advanced due to the exceptionally warm... "

\section{Full List of PE and DE ranked on the 11 unseen datasets}
\label{appen:expert_rankings}
Table \ref{table:DE_list} shows the full list of DE and Table \ref{table:PE_list} shows the full list of PE, both lists sorted in descending order with regards to the mean accuracy on 11 unseen tasks. 
\begin{table}[]
\centering
\begin{tabular}{l|c|l}
\toprule
Dataset                       & AVG         & \multicolumn{1}{c}{Categories} \\
\midrule
cosmos\_qa                    & 53.35229377 & MCQA                           \\
social\_i\_qa                 & 52.9111819  & MCQA                           \\
dream                         & 51.45885188 & MCQA                           \\
quail                         & 50.4459655  & MCQA                           \\
qasc                          & 48.05781887 & MCQA                           \\
paws/labeled\_final           & 47.65196514 & Paraph.                        \\
commonsense\_qa               & 47.20113697 & MCQA                           \\
\bottomrule
\end{tabular}
\caption{\footnotesize The full list of Dataset Experts (DE) ranked in the mean accuracy on the 11 unseen tasks. The evaluations are performed on 300 sample instances of each unseen task for efficiency.}
\label{table:DE_list}
\end{table}

\begin{table}[]
\resizebox{\textwidth}{!}{
\begin{tabular}{llcc}
\toprule
Dataset                       & Prompt                                                           & AVG         & \multicolumn{1}{c}{Task Category} \\
\midrule
cosmos\_qa                    & no\_prompt\_text                                                 & 54.65821845 & MCQA                                         \\
cosmos\_qa                    & context\_question\_description\_answer\_text                     & 54.3060466  & MCQA                                         \\
cosmos\_qa                    & context\_description\_question\_answer\_text                     & 54.19701579 & MCQA                                         \\
cosmos\_qa                    & description\_context\_question\_answer\_text                     & 53.15591518 & MCQA                                         \\
social\_i\_qa                 & Show choices and generate answer                                 & 53.06841536 & MCQA                                         \\
dream                         & baseline                                                         & 51.85164999 & MCQA                                         \\
dream                         & read\_the\_following\_conversation\_and\_answer\_the\_question   & 51.67431073 & MCQA                                         \\
\bottomrule
\end{tabular}
}
\end{table}

\begin{table}[]
\resizebox{\textwidth}{!}{
\begin{tabular}{llcc}
\toprule
Dataset                       & Prompt                                                           & AVG         & \multicolumn{1}{c}{Task Category} \\
\midrule
cos\_e/v1.11                  & explain\_why\_human                                              & 44.30354265 & MCQA                                         \\
ropes                         & background\_situation\_middle                                    & 44.21042071 & EQA                                          \\
wiqa                          & effect\_with\_string\_answer                                     & 44.16834366 & MCQA                                         \\
commonsense\_qa               & answer\_given\_question\_without\_options                        & 44.051375   & MCQA                                         \\
trec                          & pick\_the\_best\_descriptor                                      & 44.04277181 & Topic C.                                     \\
social\_i\_qa                 & Generate the question from the answer                            & 44.04212031 & MCQA                                         \\
adversarial\_qa/adversarialQA & answer\_the\_following\_q                                        & 44.03344043 & EQA                                          \\
\bottomrule
\end{tabular}
}
\end{table}

\begin{table}[]
\resizebox{\textwidth}{!}{
\begin{tabular}{llcc}
\toprule
Dataset                       & Prompt                                                           & AVG         & \multicolumn{1}{c}{Task Category} \\
\midrule
ropes                         & read\_background\_situation                                      & 43.09034626 & EQA                                          \\
ag\_news                      & classify\_with\_choices\_question\_first                         & 43.07124399 & Topic C.                                     \\
quail                         & description\_context\_question\_text                             & 43.06430021 & MCQA                                         \\
adversarial\_qa/adversarialQA & question\_context\_answer                                        & 43.04578872 & EQA                                          \\
trec                          & fine\_grained\_open\_context\_first                              & 43.04356783 & Topic C.                                     \\
dream                         & generate-first-utterance                                         & 43.04159372 & MCQA                                         \\
ropes                         & background\_new\_situation\_answer                               & 43.00629035 & EQA                                          \\
\bottomrule
\end{tabular}
}
\end{table}

\begin{table}[]
\resizebox{\textwidth}{!}{
\begin{tabular}{llcc}
\toprule
Dataset                       & Prompt                                                           & AVG         & \multicolumn{1}{c}{Task Category} \\
\midrule
xsum                          & DOC\_boils\_down\_to\_simple\_idea\_that                         & 42.38486858 & Summ.                                        \\
xsum                          & summarize\_DOC                                                   & 42.38486858 & Summ.                                        \\
xsum                          & summarize\_this\_DOC\_summary                                    & 42.38486858 & Summ.                                        \\
cosmos\_qa                    & context\_description\_question\_text                             & 42.3758318  & MCQA                                         \\
quoref                        & What Is The Answer                                               & 42.32137551 & EQA                                          \\
samsum                        & Generate a summary for this dialogue                             & 42.31155911 & Summ.                                        \\
glue/mrpc                     & want to know                                                     & 42.29343352 & Paraph.                                      \\
\bottomrule
\end{tabular}
}
\end{table}

\begin{table}[]
\resizebox{\textwidth}{!}{
\begin{tabular}{llcc}
\toprule
Dataset                       & Prompt                                                           & AVG         & \multicolumn{1}{c}{Task Category} \\
\midrule
xsum                          & DOC\_given\_above\_write\_one\_sentence                          & 41.81263384 & Summ.                                        \\
xsum                          & read\_below\_DOC\_write\_abstract                                & 41.81263384 & Summ.                                        \\
xsum                          & DOC\_tldr                                                        & 41.81263384 & Summ.                                        \\
rotten\_tomatoes              & Writer Expressed Sentiment                                       & 41.80526158 & Senti.                                       \\
imdb                          & Movie Expressed Sentiment 2                                      & 41.80336688 & Senti.                                       \\
wiki\_hop/original            & choose\_best\_object\_interrogative\_1                           & 41.78715174 & MCQA                                         \\
wiki\_hop/original            & explain\_relation                                                & 41.78715174 & MCQA                                         \\
\bottomrule
\end{tabular}
}
\end{table}

\begin{table}[ht!]
\resizebox{\textwidth}{!}{
\begin{tabular}{llcc}
\toprule
Dataset                       & Prompt                                                           & AVG         & \multicolumn{1}{c}{Task Category} \\
\midrule
duorc/ParaphraseRC            & question\_answering                                              & 41.25012619 & EQA                                          \\
duorc/ParaphraseRC            & movie\_director                                                  & 41.25012619 & EQA                                          \\
duorc/ParaphraseRC            & generate\_question                                               & 41.25012619 & EQA                                          \\
duorc/ParaphraseRC            & extract\_answer                                                  & 41.25012619 & EQA                                          \\
duorc/ParaphraseRC            & title\_generation                                                & 41.25012619 & EQA                                          \\
duorc/ParaphraseRC            & answer\_question                                                 & 41.25012619 & EQA                                          \\
duorc/ParaphraseRC            & generate\_question\_by\_answer                                   & 41.25012619 & EQA                                          \\
\bottomrule
\end{tabular}
}
\caption{\footnotesize The full list of Prompt Experts (PE) ranked in the mean accuracy on the 11 unseen tasks. The evaluations are performed on 300 sample instances of each unseen task for efficiency.}
\label{table:PE_list}
\end{table}

\end{document}